\documentclass[journal]{IEEEtran}
\usepackage[switch]{lineno}
\usepackage{hyperref}
\usepackage{amsmath,amssymb}
\usepackage{booktabs}
\usepackage{algorithm}
\usepackage{algorithmic}
\usepackage{amsthm}
\usepackage{multirow}
\usepackage{graphicx}

\newtheorem{prop}{Proposition}
\modulolinenumbers[5]
\newcommand{\mbx}{\mathbf{x}}
\ifCLASSINFOpdf
\else
\fi
\begin{document}
\title{Enhanced Few-shot Learning for Intrusion Detection in Railway Video Surveillance}

\author{Xiao~Gong,~
        Xi~Chen,~
        and~Wei~Chen,~\IEEEmembership{Senior~Member,~IEEE}% <-this % stops a space
\thanks{Xiao Gong and Wei Chen are with the State Key Laboratory of Rail Traffic Control and Safety, Beijing Jiaotong University, Beijing, China (email: xiaogong@bjtu.edu.cn, weich@bjtu.edu.cn).}
\thanks{Xi Chen is with the Department of Computer Science, University of Bath, Bath, UK (email: xc841@bath.ac.uk).}
\thanks{Equal co-first author: Xi~Chen.}
\thanks{Corresponding author: Wei Chen.}% <-this % stops a space
}

\maketitle

\begin{abstract}
Video surveillance is gaining increasing popularity to assist in railway intrusion detection in recent years. However, efficient and accurate intrusion detection remains a challenging issue due to: (a) limited sample number: only small sample size (or portion) of intrusive video frames is available; (b) low inter-scene dissimilarity: various railway track area scenes are captured by cameras installed in different landforms; (c) high intra-scene similarity: the video frames captured by an individual camera share a same backgound. In this paper, an efficient few-shot learning solution is developed to address the above issues. In particular, an enhanced model-agnostic meta-learner is trained using both the original video frames and segmented masks of track area extracted from the video. Moreover, theoretical analysis and engineering solutions are provided to cope with the highly similar video frames in the meta-model training phase. The proposed method is tested on realistic railway video dataset. Numerical results show that the enhanced meta-learner successfully adapts unseen scene with only few newly collected video frame samples, and its intrusion detection accuracy outperforms that of the standard randomly initialized supervised learning.
\end{abstract}

\begin{IEEEkeywords}
Railway intrusion detection, few-shot learning, meta-learner, video surveillance.
\end{IEEEkeywords}

% \fi
\IEEEpeerreviewmaketitle

\section{Introduction}
\begin{figure}[t!]
\centering
\includegraphics[width=0.99\columnwidth]{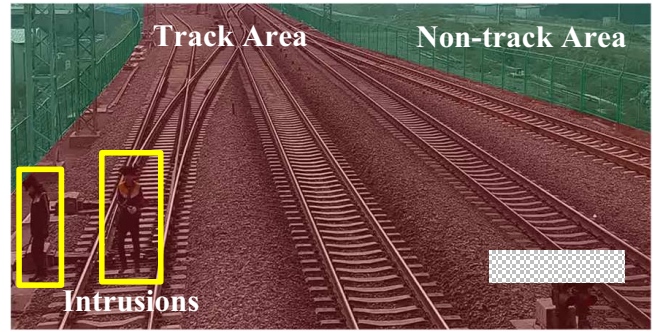}
\caption{Intruder detection in a video surveillance system.}
\vskip -0.1in
\label{fig:illuse}
\end{figure}

\IEEEPARstart{R}{ailway} intrusion detection plays an important role in railway management system and assists in safe operation of the trains with the rapid development of the railway system all over the world. Modern intrusion detection techniques has been extensively explored in the past two decades to improve the safety of both the railway users and the facilities~\cite{4730859,101177}. For example, a modern video surveillance system, as shown in Figure~\ref{fig:illuse}, is used to monitor the railway track areas and potential human intruders through video streaming. However, the unauthorised human intrusions and nature-caused track obstacles are still the main fuses of railway traffic accidents in China, thus the railway safety is paid much more attention than ever before\footnote{http://www.nra.gov.cn/jgzf/zfjg/zfdt/202003/t20200327\_107025.shtml}.

%http://www.nra.gov.cn/jgzf/zfjg/zfdt/202003/t20200327_107025.shtml

Railway intrusion detection methods can be broadly classified into two main categories, namely, contact and non-contact detection~\cite{guo2012intrusion}. The former type of methods employ tools such as protective net, of which the physical deformation caused by intrusion can be captured. One example is fiber Bragg grating, which is a type of Bragg reflector that is capable of detecting intrusions based on Bragg wavelength shift caused by human footsteps~\cite{CATALANO201791}. The latter type, non-contact detection, which often takes advantage of contactless sensing, is favoured by industrial practitioners in recent years. Successful applications include (but not limit to): infrared detection~\cite{4692395}, laser detection~\cite{luy2018initial}, and detection using machine vision~\cite{pu2014study}. Amongst the various contactless techniques, intrusion detection using vision data collected by video surveillance camera is the one that attracts more and more attentions in the wake of its wide monitoring vision angle, easy installation and convenient maintenance features.

%%%%%%%%%%%%%%%%%%%%%%%%%%%%%%%%%%%%%%%%%%%%%%%%%%%%%%%%%%%%%%%%%%%%%%%%%%%%%%%%%%%%%%%
%%%%%%%%%%%%%%%%%%%%%%%%%%%%%%%%%%%%%%%%%%%%%%%%%%%%%%%%%%%%%%%%%%%%%%%%%%%%%%%%%%%%%%%%
\par

The success of deep learning techniques in computer vision applications inspires one to explore the potential of applying deep neural nets in railway video monitoring scenario. However, typical deep architecture in computer vision requires a considerable amount of labelled training data for the neural nets to learn the latent data distribution and to avoid overfitting~\cite{NIPS2012_4824}. Besides, three practical barriers exist that prevent one from exploring the technique in railway monitoring applications. Firstly, the ratio between non-intrusive and intrusive video frames is relatively large. In other words, only a small number of video frames contain intrusion events, while the rest remain unchanged. Secondly, practical video surveillance solutions often need to install hundreds and thousands of fixed pan-tilt-zoom (PTZ) video cameras along the railway tracks to cover as much monitoring areas as possible. New video streams with completely unseen landform are highly likely to appear out-of-sample, which can results in failure of intruder detection or frequent false alarms. Thirdly, the `effective' training samples from an individual video camera can be very limited, as most of the video frames are highly similar under a same track scene. This clearly leads to a computer vision scenario with small samples. Moreover, data collection and labelling can be a time consuming and labour intensive task in railway applications. Thus, efficient intrusion detection using small sample size is of interests to a wide range of practitioners.

In this paper, an efficient video intrusion detection algorithm using deep learning techniques is developed. A few-shot meta-model of deep convolutional neural networks (CNN) is trained by incorporating the idea of model-agnostic meta-learning (MAML) algorithm~\cite{MAML}. The meta-model is further enhanced using a feature engineering scheme to improve algorithm performance. As far as we know, this is the first time an approach of this type is applied to solve the small sample problem in the smart railway video surveillance scenario. The contributions of this work are summarised as follows:
\begin{itemize}
\item An novel enhanced few-shot learning framework is introduced, which could be easily applied to other railway video intrusion applications. The algorithmic components of the framework can be replaced by alternative user chosen algorithms (or networks) without affecting the effectiveness of the whole enhanced few-shot learning pipeline.

\item The trained meta-model requires only a few of unseen training samples and gradient descent steps to be adaptive to a new scene, which can be economically and computationally efficient in practical deployments.

\item A feature engineering scheme is developed to further improve the prediction accuracy of the trained meta-learner. This is conducted by extracting rail track features using a pre-tuned fully convolutional networks (FCN)~\cite{7478072}.

\item A theoretical proof is provided to analyse the relationship between the highly similar video frames and the overfitting phenomenon observed in the meta-model training phase. Effective engineering solutions are supplied to address the high intra-scene similarity issue to prevent the meta-model from overfitting.
\end{itemize}

\section{Related work}
\subsection{Vision-based railway intrusion detection}

Vision-based intrusion detection has seen a rapid progress in recent years. Some lately developed video detection applications were proposed using existing techniques, such as principal components analysis (PCA)~\cite{1336499}, Kalman filtering~\cite{Shi2015Study} and frame difference~\cite{Guo2016High,nakasone2017frontal}. One also sees blooming of CNN research in computer vision area as a powerful feature extraction tool. CNN now has been an important technique in smart railway and has been successfully applied in many railway research work~\cite{7506117,8006280,8516370}. Meanwhile, deep learning techniques brought in new ideas to the railway intrusion detection research field. For instances, Wang \emph{et al.}~\cite{wang2019adaptive} proposed a segmentation algorithm which utilised a simple designed CNN to classify all local areas of input image using a set of Gaussian kernels.  Although this method has relatively higher runtime efficiency in GPU-free computing environment, it requires an extra classification module to perform the intrusion detection task. Huang \emph{et al.}~\cite{8832957} employed an improved VGG-Net~\cite{simonyan2014very} to perform intrusion detection by training a classifier using aerial video images. However, collecting samples using the unmanned aerial vehicle (UAV) is a relatively expensive way than video cameras to maintain surveilliance. Guo \emph{et al.}~\cite{guo2019high} used a single shot multibox detector (SSD)~\cite{liu2016ssd} to detect and bound the rail intrusions, where a deconvolution structure was introduced into the SSD to improve the detection ability of small objects. Combining the advantages of two state-of-the-art detectors, i.e., the Faster R-CNN~\cite{7485869} and SSD~\cite{liu2016ssd}, Ye \emph{et al.}~\cite{8978612} designed an efficient differential feature fusion CNN to detect obstacles on railway tracks. However, all the above methods require a considerable number of training samples to be collected and processed beforehand. As far as we know, no methodology has been proposed to solve the railway intrusion detection problem with a small sample size.

\subsection{Meta-learning}
% transfer meta-knowledge onto initialization
Meta-learning, or learning to learn~\cite{schmidhuber}, aims to acquire inductive bias that can describes features of the entire task pool where each individual task has its own train/test sets. A wide range of research has been done in meta-learning field to extract `transferable knowledge' from the training tasks~\cite{MAML,NIPS2016_6385,NIPS2017_6996,Sung_2018_CVPR,RaviL17}. An important variant of meta-learning algorithms is to train meta-models for unseen tasks with very limited samples (or supervision), also know as \emph{few-shot learning}.

Among various meta-learning approaches, MAML is one of the powerful framework that has been successfully applied to solve the few-shot learning problem in many disciplines, such as compress sensing~\cite{wu19d}, internet of things (IoT)~\cite{8815426} and telecommunications~\cite{8761319}. Compared with classical pre-training methods that may require relatively slower pre-tunning of model parameters, MAML aims to quickly learn suitable initial condition set using gradient decent. The meta-learner is model-agnostic in the sense that it can be directly applied to any learning problem and model optimised using gradient decent.

\section{Problem formulation}

%%%%%%%%%%%%%%%%%%%%%%%%%%%%%%%%%%%%%%%%%%%%%%%%%%%%%%%%%%%%%%%%%%%%%%%%%%%%%%%%%%%%%%%
\begin{figure}[t!]
\centering
\includegraphics[width=0.99\columnwidth]{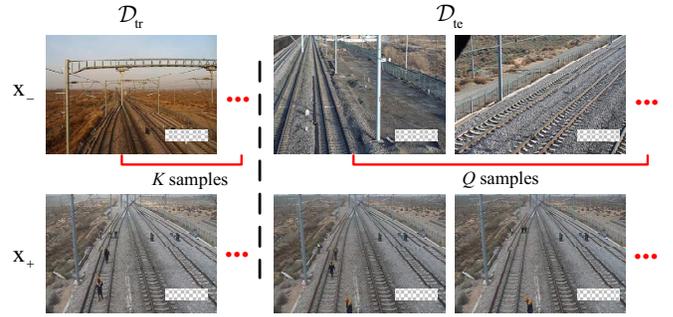}
\caption{An example of the $K$-shot intrusion detection task, where $\mbx_-$ and $\mbx_+$ denote the samples of non-intrusion and intrusion scenes, respectively. There are $K$ and $Q$ samples in each scene for training set $\mathcal{D}_{\text{tr}}$ and test set $\mathcal{D}_{\text{te}}$, respectively. }
%\vskip -0.1in
\label{fig:illutask}
\end{figure}
%%%%%%%%%%%%%%%%%%%%%%%%%%%%%%%%%%%%%%%%%%%%%%%%%%%%%%%%%%%%%%%%%%%%%%%%%%%%%%%%%%%%%%%%
In the railway intruder detection example, one is interested in the occurrence of true intrusion events. From a machine learning perspective, this motivating application can be cast as a binary classification problem where the binary valued intrusion occurrence is predicted in light of video camera streaming data.

Consider a distribution over tasks $ p(\mathcal{T})$. A task $\mathcal{T}_i$ drawn from $ p(\mathcal{T})$ is a $K$-shot learning problem where $K$ intrusive and $K$ non-intrusive video frames are available for model training, i.e., $2K$ training samples in total. We also denote $Q$ as the testing sample number, with $Q$ samples for both intrusive and non-intrusive scenes. As shown in Figure~\ref{fig:illutask}, we denote the full data set as $\{\mathcal{D}^{(i)}_{\text{tr}},\mathcal{D}^{(i)}_{\text{te}}\}$, where the former represents training set and the latter represents testing set. Note that $\mathcal{T}_i$ is a standard supervised learning task that consists training set $\mathcal{D}^{(i)}_{\text{tr}}$ and test set $\mathcal{D}^{(i)}_{\text{te}}$. In the meta-learning settings, $\mathcal{D}^{(i)}_{\text{tr}}$ and $\mathcal{D}^{(i)}_{\text{te}}$ are also called \emph{support set} and \emph{query set}, respectively. For the $i$th task, both the data set can be expressed in the form as:
%%%%%%%%%%%%%%%%%%%%%%%%%%%%%%%%%%%%%%%%%%%%%%%%%%%%%%%%%%%%%%%%%
\begin{align*}
    \mathcal{D}^{(i)} = \{\mbx^{(i)}_{+,k},y^{(i)}_{+,k},\mbx^{(i)}_{-,k},y^{(i)}_{-,k}\}^{K}_{k=1},
\end{align*}
%%%%%%%%%%%%%%%%%%%%%%%%%%%%%%%%%%%%%%%%%%%%%%%%%%%%%%%%%%%%%%%%%
where $\mbx$ denotes the input (tensor) that represents the raw video frames and $y$ denotes the corresponding one-hot label (binary event occurance). Symbols $+$ and $-$ denote intrusion and non-intrusion scenes.

In practice, the $i$th classification task can be trained by its corresponding classifier $f_{\theta}^{(i)}$, where $\theta$ represents its model parameters. An essential step of meta-learning approaches is to learn initial condition (denoted as $\theta_0$) of the parameter set $\theta$ for \emph{base model} (the meta-learner). This can be achieved by firstly updating $\theta^{(i)}$ in each classifier $f_{\theta}^{(i)}(\cdot)$ by standard gradient descent as follows
%%%%%%%%%%%%%%%%%%%%%%%%%%%%%%%%%%%%%%%%%%%%%%%%%%%%%%%%%%%%%%%%%
\begin{equation}
\begin{split}
\theta^{(i)}_n = \theta^{(i)}_{n-1} - \alpha\nabla_{\theta^{(i)}_{n-1}}\mathcal{L}_{\text{tr}}(f_{\theta_{n-1}}^{(i)}),
\end{split}
\label{innerup}
\end{equation}
%%%%%%%%%%%%%%%%%%%%%%%%%%%%%%%%%%%%%%%%%%%%%%%%%%%%%%%%%%%%%%%%%
where $n = 1,\ldots, N$ denotes the recursive updating steps, $\mathcal{L}_{\text{tr}}(\cdot)$ represents the loss function on the support set, and $\alpha$ is a coefficient describes inner-loop learning rate.

Once the task-wise parameter sets $\{\theta^{(i)}\}_{i=1}^{I}$ of $I$ tasks have been updated, the meta-update of the initial condition $\theta_0$ can be computed by
%%%%%%%%%%%%%%%%%%%%%%%%%%%%%%%%%%%%%%%%%%%%%%%%%%%%%%%%%%%%%%%%%%
\begin{equation}
\begin{split}
\theta_0 \leftarrow \theta_0 - \beta\sum_{i=1}^{I}\nabla_{\theta_0}\mathcal{L}_{\text{te}}(f^{(i)}_{\theta_N}),
\end{split}
\label{metaup}
\end{equation}
%%%%%%%%%%%%%%%%%%%%%%%%%%%%%%%%%%%%%%%%%%%%%%%%%%%%%%%%%%%%%%%%%%
where $\mathcal{L}_{\text{te}}(\cdot)$ denotes the loss function on the query set, and $I$ denotes the meta-batch size and $\beta$ is the coefficient describes the meta-learning rate.

The gradient of loss function in Equation \eqref{metaup} can be further expanded by substitution using Equation \eqref{innerup}, as
%%%%%%%%%%%%%%%%%%%%%%%%%%%%%%%%%%%%%%%%%%%%%%%%%%%%%%%%%%%%%%%%%%
\begin{align*}
&\nabla_{\theta_0}\mathcal{L}_{\text{te}}(f^{(i)}_{\theta_N}) \\
&=\nabla_{\theta^{(i)}_N}\mathcal{L}_{\text{te}}(f^{(i)}_{\theta_N})\prod_{n=1}^N\nabla_{\theta^{(i)}_{n-1}}\left(\theta^{(i)}_{n-1}-\alpha\nabla_{\theta^{(i)}_{n-1}}\mathcal{L}_{\text{tr}}(f^{(i)}_{\theta_{n-1}})\right).
\label{secgd}
\end{align*}
%%%%%%%%%%%%%%%%%%%%%%%%%%%%%%%%%%%%%%%%%%%%%%%%%%%%%%%%%%%%%%%%%
The second derivatives on the right hand side of the equation can be omitted (see~\cite{MAML} for details), and give a simplified Equation \eqref{metaup}, rewritten as:
%%%%%%%%%%%%%%%%%%%%%%%%%%%%%%%%%%%%%%%%%%%%%%%%%%%%%%%%%%%%%%%%%%
\begin{equation}
\begin{split}
\theta_0 \leftarrow \theta_0 - \beta\sum_{i=1}^{I}\nabla_{\theta^{(i)}_N}\mathcal{L}_{\text{te}}(f^{(i)}_{\theta_N}),
\end{split}
\label{firmeta}
\end{equation}
%%%%%%%%%%%%%%%%%%%%%%%%%%%%%%%%%%%%%%%%%%%%%%%%%%%%%%%%%%%%%%%%%%
which significantly reduces computational expense but can achieve comparable performance as that in the standard MAML~\cite{MAML}. In this paper, we employ this simplified meta-learner as the basis of the proposed few-shot learning algorithm. Given the well tuned initial condition $\theta_0$ of the base model, the meta-learner can now quickly adaptive to the unseen scene with $2K$ new samples.

\section{Few-shot learning for intrusion detection}
The standard MAML framework has demonstrated its performance in various few-shot learning applications, which makes it a promising candidate for the railway video intrusion detection problem. However, before applying the powerful framework, some issues as identified earlier, i.e., low inter-scene dissimilarity and high intra-scene similarity, must be addressed.

To facilitate understanding in this paper, the similarity between two images corresponds to the pixel-wise difference. In addition, we denote $\text{Sim}(\mathcal{D}^{(i)}, \mathcal{D}^{(j)})$ as the similarity between two tasks which is formulated by
%%%%%%%%%%%%%%%%%%%%%%%%%%%%%%%%%%%%%%%%%%%%%%%%%%%%%%%%%%%%%%%%%%
\begin{equation}
\small{\begin{split}
&\text{Sim}(\mathcal{D}^{(i)}, \mathcal{D}^{(j)}) \\
=
&\frac{1}{2K\sqrt{N}}\left(\left \|\sum_{k=1}^{K}\left(\bar{\mbx}^{(i)}_{+,k} -\bar{\mbx}^{(j)}_{+,k} \right)\right \|_2
+\left \|\sum_{k=1}^{K}\left(\bar{\mbx}^{(i)}_{-,k} -\bar{\mbx}^{(j)}_{-,k} \right)\right \|_2\right),
\end{split}}
\label{tasksim}
\end{equation}
%%%%%%%%%%%%%%%%%%%%%%%%%%%%%%%%%%%%%%%%%%%%%%%%%%%%%%%%%%%%%%%%%%
where $\bar{\mbx}$ is a vector that denotes the flattened image $\mbx$ and $N$ denotes the length of $\bar{\mbx}$. The smaller the value of $\text{Sim}(\mathcal{D}^{(i)}, \mathcal{D}^{(j)})$, the higher the similarity between $\mathcal{D}^{(i)}$ and $\mathcal{D}^{(j)}$.

\subsection{The datasets}
This subsection details the datasets used in this paper. We compare the difference of both the local railway images dataset captured by video cameras and a bench-marking Minimagenet dataset used in the research community.
%%%%%%%%%%%%%%%%%%%%%%%%%%%%%%%%%%%%%%%%%%%%%%%%%%%%%%%%%%%%%%%%%%
\begin{figure}[t!]
\centering
\includegraphics[width=0.99\columnwidth]{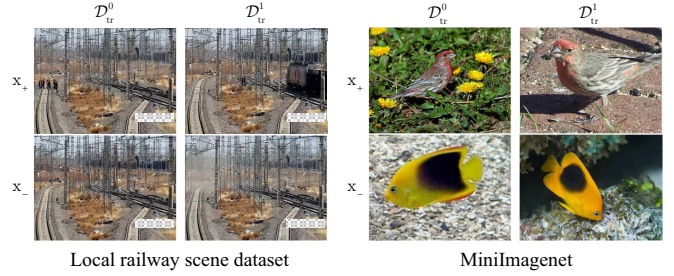}
%\vskip -0.1in
\caption{Examples of two datasets used in this paper. Examples on the
left panel were from the railway scene dataset collected in China railway.
The right panel examples were from the bench-marking MiniImagenet dataset.  $\mbx_+$ and $\mbx_-$ denote the positive and negative samples, respectively.}
\label{fig:simtask1}
%\vskip -0.05in
\end{figure}
%%%%%%%%%%%%%%%%%%%%%%%%%%%%%%%%%%%%%%%%%%%%%%%%%%%%%%%%%%%%%%%%%%
%%%%%%%%%%%%%%%%%%%%%%%%%%%%%%%%%%%%%%%%%%%%%%%%%%%%%%%%%%%%%%%%%%
\begin{figure}[t!]
\centering
\includegraphics[width=0.99\columnwidth]{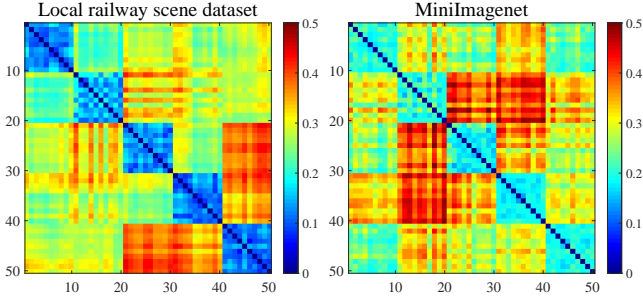}
%\vskip -0.1in
\caption{Cross-similarity of all $50$ tasks constructed from $5$ scenes in the local railway scene dataset (left panel) and $5$ types of binary classification tasks in the MiniImagenet dataset (right panel).}
\label{fig:simtask2}
\vskip -0.1in
\end{figure}
%%%%%%%%%%%%%%%%%%%%%%%%%%%%%%%%%%%%%%%%%%%%%%%%%%%%%%%%%%%%%%%%%%

Example images of the railway dataset are shown in the left panel of Figure~\ref{fig:simtask1}. This dataset was collected at railway `throat areas', where a large number of turnouts and crossings exist. This typical `busy' railway track region often requires intensive worker operations and maintenance, and therefore accurate intrusion detection and monitoring are desired. Particularly, the video frames data was collected from PTZ cameras deployed at $65$ local throat areas along the railway tracks between December $2018$ and April $2019$ in China. The raw frames were cleaned by standard pre-processing steps to extract valid images and to remove the majority of similar frames in each scene. Specifically, we employed standard normalised cross correlation, mean square error (MSE) and structural similarity (SSIM) metrics to evaluate the eligibility of the frames and thus filter out highly identical frames.

The `small samples' issue also significantly affected the whole data collection process. Although an intruder event, once occurred, would greatly threaten railway passenger safety, it seldomly happened on a daily basis. In fact, some of the throat area scenes only contained a small portion of eligible frames for the training phase. To secure sufficient number of training samples in each of the meta-learning tasks, we fused the $65$ subsets of eligible frames into $22$ groups of scenes, i.e., each group was composed by $2$ to $3$ subsets frames. The processed scenes were then divided into three separate portions: $18$ scenes were used for meta-training phase, $2$ scenes for meta-cross-validation phase, and the rest $2$ scenes for meta-test phase. The whole dataset contained $11409$ RGB images, which was composed by $5863$ intrusion and $5546$ non-intrusion images. An image would be manually labelled as `intrusion' if workers or big animals (e.g. dog) appeared in the track area. All images were resized in a same $640\times 480$ resolution.

The Minimagenet dataset~\cite{RaviL17}, which includes $64$ training classes, is illustrated on the right panel of Figure~\ref{fig:simtask1}. There are $\frac{64\times 63}{2}$ types of binary classification tasks by permutation for training of the meta-learning models (see~\cite{RaviL17} for detailed description of the Minimagenet dataset). A comparison worth to note is that the MiniImagenet dataset was `better' constructed when compared to the railway dataset. In particular, images from the MiniImagenet dataset have much lower intra-similarity (where images within the same class are differ from each other), whereas railway data frames in a same scene have significantly high similarity as they share a same background scene. To verify that the MiniImagenet dataset is better constructed, we randomly select $5$ types of binary classification tasks from the MiniImagenet dataset and $5$ scenes from the railway dataset, we then draw $10$ tasks with $K=10$ from each type/scene. The cross-similarity of all $50$ tasks is shown in Figure~\ref{fig:simtask2}, where one sees that the railway scene tasks has low inter-scene dissimilarity and higher intra-scene similarity when compared to those of the MiniImagenet dataset.

\subsection{An improved base model}
One special feature of the railway video camera dataset is its high similarity across the video frames of a scene and its relatively high dissimilarity between video frames of different landform scenes. As demonstrated in Equation \eqref{firmeta}, the standard MAML framework updates model parameters through gradient descent. Thanks to the the low inter-type dissimilarity and low intra-type similarity in MiniImagenet, the diversity of meta-training tasks and meta-testing tasks are relatively balanced which is helpful to reduce overfitting. However, the tasks of different local railway scenes are unbalanced due to the low inter-scene dissimilarity and higher intra-scene similarity, which may easily lead to overfitting of the meta-model under the special features of railway video frames.

\begin{figure}[t!]
\centering
\includegraphics[width=0.99\columnwidth]{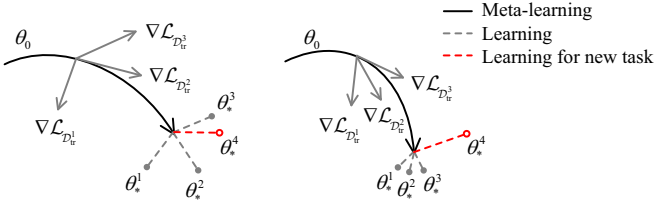}
\caption{Diagrams of standard MAML (left) that quickly adapts to a new task (in red dashed line), while in tasks with high dissimilarity, the standard MAML leads to overfitting.}
%\vskip -0.1in
\label{fig:simtask}
\end{figure}

This issue has been further illustrated in Figure~\ref{fig:simtask},
where samples in each railway video task contribute very similar gradient directions which makes the meta-learner harder to quickly adapt to the sample in an unseen task (the right figure in Figure~\ref{fig:simtask}). An even worse condition is that increasing the sample size of a scene in the training set merely improve the diversity of the gradient directions.

In this paper, the relationship between the highly similar gradients and highly similar training samples has also been analysed theoretically and described in Proposition~\ref{Theorem:1}.

\par
\begin{prop}
\label{Theorem:1}
Without loss of generality, consider a base model of standard CNN that has a convolution layer with a filter $\omega$ followed by a linear layer with weight coefficients $\eta = [\eta_0,\eta_1]$, a softmax layer and a binary cross-entropy loss $\mathcal{L}$. Given two training tasks $\mathcal{D}^0 = \{\mbx^0_{+,k},y^0_{+,k}, \mbx^0_{-,k},y^0_{-,k}, \}^{K}_{k=1}$ and $\mathcal{D}^1 = \{\mbx^1_{+,k},y^1_{+,k},\mbx^1_{-,k},y^1_{-,k}\}^{K}_{k=1}$, the difference of the gradients can be approximated by the following equations:
%%%%%%%%%%%%%%%%%%%%%%%%%%%%%%%%%%%%%%%%%%%%%%%%%%%%%%%%%%%%%%%%%%
\begin{align}
\label{eq:Theorem1}
&\frac{\partial(\mathcal{L}_{\mathcal{D}^0} - \mathcal{L}_{\mathcal{D}^1})}{\partial \eta_0} \nonumber\\
&\rightarrow \frac{1}{2K}\sum_{k=1}^{K}\left(D_{\omega}\left(\bar{\mbx}^1_{+,k} -\bar{\mbx}^0_{+,k} \right) + D_{\omega}\left(\bar{\mbx}^0_{-,k} - \bar{\mbx}^1_{-,k}\right)\right),\\
\label{eq:Theorem2}
&\frac{\partial(\mathcal{L}_{\mathcal{D}^0} - \mathcal{L}_{\mathcal{D}^1})}{\partial \eta_1} \nonumber\\
&\rightarrow\frac{1}{2K}\sum_{k=1}^{K}\left(D_{\omega}\left(\bar{\mbx}^0_{+,k} - \bar{\mbx}^1_{+,k}\right) + D_{\omega}\left( \bar{\mbx}^1_{-,k}-\bar{\mbx}^0_{-,k}\right)\right),
\\
\label{eq:Theorem3}
&\frac{\partial(\mathcal{L}_{\mathcal{D}^0} - \mathcal{L}_{\mathcal{D}^1})}{\partial \bar{\omega}} \nonumber\\
&\rightarrow\frac{1}{2K}\sum_{k=1}^{K}\left(D_{\eta_0}\left( \bar{\mbx}^1_{+,k} - \bar{\mbx}^0_{+,k}\right) + D_{\eta_1}\left(\bar{\mbx}^1_{-,k}-\bar{\mbx}^0_{-,k} \right)\right),
\end{align}
%%%%%%%%%%%%%%%%%%%%%%%%%%%%%%%%%%%%%%%%%%%%%%%%%%%%%%%%%%%%%%%%%%
where $\bar{\omega}$ and $\bar{\mbx}$ denote the identities after the flattening operation of CNN. Functions $D_{\omega}(\cdot)$, $D_{\eta_0}(\cdot)$ and $D_{\eta_1}(\cdot)$ represent the sparse Toeplitz matrices with respect to $\omega$, $\eta_0$ and $\eta_1$, respectively.
\end{prop}

%%%%%%%%%%%%%%%%%%%%%%%%%%%%%%%%%%%%%%%%%%%%%%%%%%%%%%%%%%%%%%%%%%
%%%%%%%%%%%%%%%%%%%%%%%%%%%%%%%%%%%%%%%%%%%%%%%%%%%%%%%%%%%%%%%%%%

\begin{proof}
Assume sample input $\mbx$ is an $L\times L$ matrix, the forward propagation of the base model can be expressed as
%%%%%%%%%%%%%%%%%%%%%%%%%%%%%%%%%%%%%%%%%%%%%%%%%%%%%%%%%%%%%%%%%%
\begin{gather}
\label{eq:conv} z_{ l_1l_2} = \sum_{a = 0}^{M-1}\sum_{b = 0}^{M-1}\omega_{ab}\mbx_{ (l_1 + a)(l_2 + b)},\\
%\label{eq:conv}
\label{eq:linear} u = [u_{ 0}, u_{ 1}] =\eta^T\bar{z} =[\eta_0, \eta_1]^T \bar{z},\\
\label{eq:softmax} p = [p_{ 0}, p_{ 1}] = \text{softmax}(u),
\end{gather}
%%%%%%%%%%%%%%%%%%%%%%%%%%%%%%%%%%%%%%%%%%%%%%%%%%%%%%%%%%%%%%%%%%
where $\omega$, $z$, $\bar{z}$, $\eta$, $u$ and $p$ denote an $M\times M$ filter of convolutional layer, $(L-M+1)\times (L-M+1)$ feature map, $(L-M+1)^2\times 1$ flattened feature map, $(L-M+1)^2\times 2$ weight of linear layer, $2\times 1$ output of linear layer and $2\times 1$ output of softmax layer, respectively. The binary cross-entropy loss is given by
%%%%%%%%%%%%%%%%%%%%%%%%%%%%%%%%%%%%%%%%%%%%%%%%%%%%%%%%%%%%%%%%%%
\begin{gather}
\mathcal{L}_{\mbx} = -y_{ 0}\log p_{ 0} - y_{ 1}\log p_{ 1},
\label{eq:loss}
\end{gather}
%%%%%%%%%%%%%%%%%%%%%%%%%%%%%%%%%%%%%%%%%%%%%%%%%%%%%%%%%%%%%%%%%%
where $y = [y_{ 0}, y_{ 1}]$ is the one-hot label of $\mbx$.

The backward propagation of the filter and linear layer can be derived using the chain rule. The gradient component of fully-connected weights are given by
%%%%%%%%%%%%%%%%%%%%%%%%%%%%%%%%%%%%%%%%%%%%%%%%%%%%%%%%%%%%%%%%%%
\begin{gather}
\label{eq:w1} \frac{\partial\mathcal{L}_{\mbx}}{\partial \eta_0} = \left(
\frac{\partial\mathcal{L}_{\mbx}}{\partial p_{ 0}}\frac{\partial p_{ 0}}{\partial u_{0}}
+\frac{\partial\mathcal{L}_{\mbx}}{\partial p_{1}}\frac{\partial p_{1}}{\partial u_{0}}\right)\frac{\partial u_{0}}{\partial \eta_0}=\left(-y_{0}p_{1}+y_{1}p_{ 0}\right)\bar{z},\\
\label{eq:w2} \frac{\partial\mathcal{L}_{\mbx}}{\partial \eta_1} =\left(
\frac{\partial\mathcal{L}_{\mbx}}{\partial p_{0}}\frac{\partial p_{0}}{\partial u_{1}}
+\frac{\partial\mathcal{L}_{\mbx}}{\partial p_{1}}\frac{\partial p_{1}}{\partial u_{1}}\right)\frac{\partial u_{1}}{\partial \eta_1}=\left(y_{0}p_{1}-y_{1}p_{ 0}\right)\bar{z}.
\end{gather}
%%%%%%%%%%%%%%%%%%%%%%%%%%%%%%%%%%%%%%%%%%%%%%%%%%%%%%%%%%%%%%%%%%
Assume a simple even probability on the inference output of softmax layer, i.e., $p_0, p_1\rightarrow 0.5$. Substituting the convolution operation in \eqref{eq:conv} by Toeplitz matrix gives
%%%%%%%%%%%%%%%%%%%%%%%%%%%%%%%%%%%%%%%%%%%%%%%%%%%%%%%%%%%%%%%%%%
\begin{gather}
\label{eq:w1D} \frac{\partial\mathcal{L}_{\mbx}}{\partial \eta_0}
=\left(-y_{0}p_{1}+y_{1}p_{0}\right)D_{\omega}\bar{\mbx} \rightarrow \left\{
\begin{aligned}
&\!-\frac{1}{2}D_{\omega}\bar{\mbx}\ &\text{if}\  \mbx = \mbx_+\\
&\frac{1}{2}D_{\omega}\bar{\mbx}\ &\text{if}\  \mbx = \mbx_-
\end{aligned}
\right.,\\
\label{eq:w2D} \frac{\partial\mathcal{L}_{\mbx}}{\partial \eta_1}
=\left(y_{0}p_{1}-y_{1}p_{0}\right)D_{\omega}\bar{\mbx} \rightarrow \left\{
\begin{aligned}
&\frac{1}{2}D_{\omega}\bar{\mbx}\ &\text{if}\  \mbx = \mbx_+\\
&\!-\frac{1}{2}D_{\omega}\bar{\mbx}\ &\text{if}\  \mbx = \mbx_-
\end{aligned}
\right.,
\end{gather}
%%%%%%%%%%%%%%%%%%%%%%%%%%%%%%%%%%%%%%%%%%%%%%%%%%%%%%%%%%%%%%%%%%
where $D_{\omega}\in \mathbb{R}^{(L-M+1)^2\times L^2}$ denotes the Toeplitz matrix with respect to the filter $\omega$ and $\bar{\mbx}$ denotes the flattened $\mbx$. Note that $D_{\omega}$ is a sparse matrix. Hence Equation~\eqref{eq:Theorem1} and \eqref{eq:Theorem2} can be obtained by substituting $\mathcal{D}^0$ and $\mathcal{D}^1$ into Equation~\eqref{eq:w1D} and \eqref{eq:w2D}.

The gradient component of the filter in convolution layer can be written as
%%%%%%%%%%%%%%%%%%%%%%%%%%%%%%%%%%%%%%%%%%%%%%%%%%%%%%%%%%%%%%%%%%
\begin{gather}
\label{eq:omega} \frac{\partial\mathcal{L}_{\mbx}}{\partial \omega_{ab}}
= \sum_{l_1 = 0}^{L-M}\sum_{l_2 = 0}^{L-M}\frac{\partial\mathcal{L}_{\mbx}}{\partial z_{l_1l_2}}\mbx_{(l_1 + a)(l_2 + b)}.
\end{gather}
%%%%%%%%%%%%%%%%%%%%%%%%%%%%%%%%%%%%%%%%%%%%%%%%%%%%%%%%%%%%%%%%%%
We can further derive
%%%%%%%%%%%%%%%%%%%%%%%%%%%%%%%%%%%%%%%%%%%%%%%%%%%%%%%%%%%%%%%%%%
\begin{equation}
\begin{split}
\label{eq:difz}
\frac{\partial\mathcal{L}_{\mbx}}{\partial \bar{z}}
&= \frac{\partial\mathcal{L}_{\mbx}}{\partial p_{0}}\frac{\partial p_{0}}{\partial u_{0}}\frac{\partial u_{0}}{\partial \bar{z}} +
\frac{\partial\mathcal{L}_{\mbx}}{\partial p_{1}}\frac{\partial p_{1}}{\partial u_{1}}\frac{\partial u_{1}}{\partial \bar{z}} \\
&= -y_{0}p_{1}\eta_0-y_{1}p_{0}\eta_1.
\end{split}
\end{equation}
%%%%%%%%%%%%%%%%%%%%%%%%%%%%%%%%%%%%%%%%%%%%%%%%%%%%%%%%%%%%%%%%%%
And this gives
%%%%%%%%%%%%%%%%%%%%%%%%%%%%%%%%%%%%%%%%%%%%%%%%%%%%%%%%%%%%%%%%%%
\begin{gather}
\label{eq:omegaD1}
\frac{\partial\mathcal{L}_{\mbx_+}}{\partial \bar{\omega}}
= p_{1}D_{\eta_0} \bar{\mbx}_+ \rightarrow \frac{1}{2}D_{\eta_0}\bar{\mbx}_+,\\
\label{eq:omegaD2}
\frac{\partial\mathcal{L}_{\mbx_-}}{\partial \bar{\omega}}
= p_{0}D_{\eta_1} \bar{\mbx}_- \rightarrow \frac{1}{2}D_{\eta_1}\bar{\mbx}_-,
\end{gather}
%%%%%%%%%%%%%%%%%%%%%%%%%%%%%%%%%%%%%%%%%%%%%%%%%%%%%%%%%%%%%%%%%%
where $D_{\eta_0}, D_{\eta_1}\in \mathbb{R}^{M^2\times L^2}$ denote the sparse Toeplitz matrices with respect to the filter $\eta_0$ and $\eta_1$ as in Equation~\eqref{eq:difz}. Equation~\eqref{eq:Theorem3} can now be obtained by substituting $\mathcal{D}^0$ and $\mathcal{D}^1$ into Equation~\eqref{eq:omegaD1} and \eqref{eq:omegaD2}.
\end{proof}

According to Proposition~\ref{Theorem:1}, the upper bound of the $\ell_2$-norm of  gradients difference has a positive correlation with the task similarity between $\mathcal{D}^{0}$ and $\mathcal{D}^{1}$.
For instance, according to the definition of task similarity in Equation~\eqref{tasksim}, the right part of Equation~\eqref{eq:Theorem1} satisfies
%%%%%%%%%%%%%%%%%%%%%%%%%%%%%%%%%%%%%%%%%%%%%%%%%%%%%%%%%%%%%%%%%%
\begin{equation}
\small{\begin{split}
\label{eq:upperb}
&\frac{1}{2K}\left \|\sum_{k=1}^{K}\left(D_{\omega}\left(\bar{\mbx}^1_{+,k} -\bar{\mbx}^0_{+,k} \right) + D_{\omega}\left(\bar{\mbx}^0_{-,k} - \bar{\mbx}^1_{-,k}\right)\right)\right \|_2 \\
\leq&\frac{1}{2K}\left \|D_{\omega}\right \|_2\left(\left \|\sum_{k=1}^{K}\left( \bar{\mbx}^1_{+,k} -\bar{\mbx}^0_{+,k} \right)\right \|_2 + \left \|\sum_{k=1}^{K}\left(\bar{\mbx}^0_{-,k} - \bar{\mbx}^1_{-,k}\right)\right \|_2\right)\\
=&L\left \|D_{\omega}\right \|_2\text{Sim}(\mathcal{D}^{0}, \mathcal{D}^{1}).
\end{split}}
\end{equation}
%%%%%%%%%%%%%%%%%%%%%%%%%%%%%%%%%%%%%%%%%%%%%%%%%%%%%%%%%%%%%%%%%%
This clearly explains the overfitting issue observed when simply appling standard MAML onto the local railway dataset, as the difference of intrusive and non-intrusive images is a sparse matrix where non-zero elements only exist in a few pixels. The sparse difference over input image samples greatly affects the diversity of the derived gradients, and therefore results in the high risk of overfitting of the trained meta-model.

We now propose two established measures to mitigate the overfitting problem of standard MAML applied in the few-shot railway dataset, namely, shuffling labels and neuron dropout.

\par
\textbf{Shuffling labels}. Conventional binary detection problem often assigns intrusion as label $1$ and non-intrusion as label $0$. This may result in overfitting of the base model in meta learning, and the issue is termed \emph{memorisation problem}~\cite{Yin20}. A studied solution is to shuffle the labels after drawing tasks from the distribution $p(\mathcal{T})$, i.e. intrusive or non-intrusive images are randomly labelled by $y = [1, 0]$ or $y = [0, 1]$, each with probability $0.5$. The technique ensures that the task-specific model cannot be inferred from the fixed class-to-label settings. The effectiveness of shuffling labels can be well explained according to Proposition~\ref{Theorem:1}.

For instance, shuffling labels makes Equation~\eqref{eq:Theorem3} be converted (with probability $0.5$) to
%%%%%%%%%%%%%%%%%%%%%%%%%%%%%%%%%%%%%%%%%%%%%%%%%%%%%%%%%%%%%%%%%%
\begin{align*}
&\frac{\partial(\mathcal{L}_{\mathcal{D}^0} - \mathcal{L}_{\mathcal{D}^1})}{\partial \bar{\omega}}  \\
&\rightarrow\frac{1}{2K}\sum_{k=1}^{K}\left(D_{\eta_0}\left( \bar{\mbx}^1_{+,k} - \bar{\mbx}^0_{-,k}\right) + D_{\eta_1}\left(\bar{\mbx}^1_{-,k}-\bar{\mbx}^0_{+,k} \right)\right),
\end{align*}
%%%%%%%%%%%%%%%%%%%%%%%%%%%%%%%%%%%%%%%%%%%%%%%%%%%%%%%%%%%%%%%%%%
which reduces the occurrence frequency zeros after subtraction of the background image samples while increases the diversity of differences in gradient directions.

\par

\textbf{Neuron dropout}. It has been shown that overfitting issue can be effectively mitigated by adding dropout layers in CNN~\cite{SrivastavaHKSS14}. Unlike traditional supervised learning where only training phase and test phase need to be evaluated, there exists four phases to assess the state of task-specific dropout layer, these includes training phase of meta-training, test phase of meta-training, training phase of meta-test and test phase of meta-test. We add a dropout layer, which will only be activated in the training phase of meta-training, for the linear layer of CNN. According to Proposition~\ref{Theorem:1}, adding the dropout layer forces Equation~\eqref{eq:Theorem1} to be converted to
%%%%%%%%%%%%%%%%%%%%%%%%%%%%%%%%%%%%%%%%%%%%%%%%%%%%%%%%%%%%%%%%%%
\begin{equation}
\small{\begin{split}
&\frac{\partial (\mathcal{L}_{\mathcal{D}^0} - \mathcal{L}_{\mathcal{D}^1})}{\partial \eta_0} \\
&\rightarrow\frac{1}{2K}\sum_{k=1}^{K}\left(d^1 \circ D_{\omega}\left(\bar{\mbx}^1_{+,k}-\bar{\mbx}^1_{-,k} \right)
-d^0 \circ D_{\omega}\left(\bar{\mbx}^0_{+,k}  - \bar{\mbx}^0_{-,k} \right)\right),
\end{split}}
\label{eq:dropout}
\end{equation}
%%%%%%%%%%%%%%%%%%%%%%%%%%%%%%%%%%%%%%%%%%%%%%%%%%%%%%%%%%%%%%%%%%
where symbol $\circ$ denotes the element-wise product and $d^0$ and $d^1$ are sparse vectors with the zero elements refer to the dropout neurons. The randomness brought in by dropout layer also effectively reduces the sparsity of the subtracted matrix as well as increases the diversity of gradient direction differences.

Another issue is that the model of CNN used in the meta-learning research community is designed for the MiniImagenet images of size $84\times84$, which is not suitable for the railway images of size $640\times 480$. Following the principle of not using more network neurons, which will not lead to more serious overfitting, we slightly modified the CNN structure. We report and compare the settings of the referenced CNN in~\cite{RaviL17} and the CNN in our proposed algorithm in Table~\ref{tab:model}.

%%%%%%%%%%%%%%%%%%%%%%%%%%%%%%%%%%%%%%%%%%%%%%%%%%%%%%%%%%%%%%%%%%
\begin{figure*}[ht!]
\centering
\includegraphics[width=1.75\columnwidth]{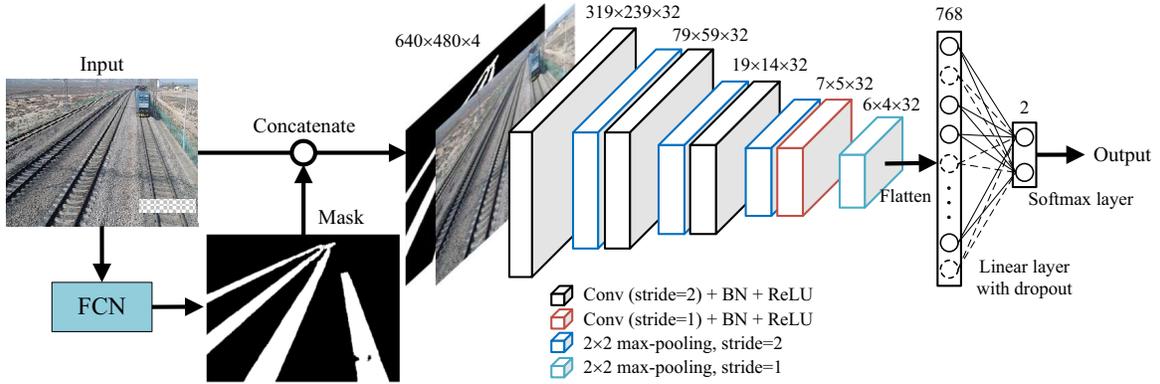}
\caption{A framework of the proposed few-shot learning algorithm in the railway track scenes. The left half of the figure describes the pre-processing and feature enhanced steps of the algorithm whereas the right half shows the detailed CNN base model architecture of the proposed meta-learner. The cuboids denote layers in the CNN, where Conv represents the convolutional layer, with batch-normalisation (BN)~\cite{ioffe2015batch} and ReLU activation. A dropout layer~\cite{SrivastavaHKSS14} is placed onto the linear layer of the CNN. The numbers above the cuboid denote the input tensor of each layer.}
%\vskip -0.1in
\label{fig:basemodel}
\end{figure*}
%%%%%%%%%%%%%%%%%%%%%%%%%%%%%%%%%%%%%%%%%%%%%%%%%%%%%%%%%%%%%%%%%%

%%%%%%%%%%%%%%%%%%%%%%%%%%%%%%%%%%%%%%%%%%%%%%%%%%%%%%%%%%%%%%%%%%
\begin{algorithm}[tb]
\caption{Enhanced few-shot learning algorithm}
\label{alg:MAML}
\textbf{Pre-requisition}: railway scene dataset, pre-defined classifiers $f_{\theta}^{(i)}$, pre-defined step size $\alpha$ and $\beta$, meta-batch size $I$ and max iteration number $J$
\begin{algorithmic}[1]
   \STATE Randomly initialise $\theta=\theta_0$
   \STATE Draw tasks from meta scenes with cross validation.
   \FOR{$j=1:J$}
   \STATE Randomly draw $I$ tasks from meta-training scenes
   \FOR{$i=1:I$}
   \STATE Recursively compute the hyper-parameter sets of each classifier via Equation~\eqref{innerup}
   \STATE Obtain $\theta^{(i)}_N$ after $N$ step iterations
   \ENDFOR
   \STATE Update parameter set of base model $\theta_0$ via Equation~\eqref{firmeta}
    \IF{converge on validation tasks}
    \STATE \textbf{break}.
    \ENDIF
   \ENDFOR
\end{algorithmic}
\end{algorithm}
%\vskip -0.2in
%%%%%%%%%%%%%%%%%%%%%%%%%%%%%%%%%%%%%%%%%%%%%%%%%%%%%%%%%%%%%%%%%%

%%%%%%%%%%%%%%%%%%%%%%%%%%%%%%%%%%%%%%%%%%%%%%%%%%%%%%%%%%%%%%%%%%
%\setlength{\tabcolsep}{3pt}
\begin{table*}[!ht]
%\small
\renewcommand\arraystretch{1}
\begin{center}
\caption{CNN model settings between the referenced model and the proposed model in this paper. Module $1$ to $4$ denote the convolutional blocks and Module $5$ denotes the linear layer, as illustrated in Figure~\ref{fig:basemodel}. The referenced CNN structure were firstly proposed in~\cite{RaviL17}.}
%\vskip 0.1in
\label{tab:model}
\begin{tabular}{|c|c|c|}
\hline
\textbf{Structure} & \textbf{Referenced CNN $f_{\theta}^0$}  & \textbf{Proposed CNN $f_{\theta}^1$} \\
\hline
Input size & $84\times84$ &$640\times480$ \\
\hline
Module $1$ & $32$ filters ($3\times3\times3$), stride $=1$ & $32$ filters ($3\times3\times3$), stride $=2$ \\
\hline
Module $2$  &$32$ filters ($3\times3\times32$), stride $=1$ &$32$ filters ($3\times3\times32$), stride $=2$ \\
\hline
Module $3$ & $32$ filters ($3\times3\times32$), stride $=1$ & $32$ filters ($3\times3\times32$), stride $=2$ \\
\hline
Module $4$ &$32$ filters ($3\times3\times32$), stride $=1$ &$32$ filters ($3\times3\times32$), stride $=1$ \\
\hline
Module $5$ &  $800\times2$ fully-connected nodes & $768\times2$ fully-connected nodes \\
\hline
Number of parameters & $32901$ & $30434$ \\
\hline
\end{tabular}
\end{center}
%\vskip -0.1in
\end{table*}
%%%%%%%%%%%%%%%%%%%%%%%%%%%%%%%%%%%%%%%%%%%%%%%%%%%%%%%%%%%%%%%%%%%%%%%%%

\subsection{Few-shot meta-learner enhanced by mask inputs}

In addition to \emph{shuffling labels} and \emph{neuron dropout}, the proposed meta-learner is further enhanced by additional mask inputs that are extracted from the original video frames (of the training data). The automatically identified mask inputs highlight the essential spatial features of track areas in each scene, and therefore assist in guiding the meta-learner to perform more accurate classification.

In practice, the mask inputs are extracted by introducing a fully convolutional network (FCN) prior to the base model training phase. Unlike classical CNN, an FCN~\cite{7478072} transforms the height and width of the intermediate layer feature map back to the size of input image through the transposed convolution layer, so that the predictions have a one-to-one correspondence with input image in spatial dimension (height and width). Being a well studied neural network model, the FCN can be straightforwardly trained offline to achieve reasonably good performance. The input of the FCN are the same video frames of the whole training set and the output are the pre-labelled true masks of the corresponding frames. In fact, the training data of the FCN doesn't necessary to be the same as those in the few-shot learning phase. A fairly larger size of data set can be used for this separate FCN training process to improve its generalisation ability and prediction performance. In this paper, we fed the same training set to both the meta-learner and the FCN.

The detailed design of the proposed few-shot meta-learning algorithm is shown in Figure~\ref{fig:basemodel}. The procedure is described as follows:
\begin{enumerate}
    \item The raw video frames (images) served as the input of a pre-trained FCN to extract the corresponding masks.
    \item The masks are then concatenated as the $4$th channel following the RGB channels of the raw video frames to compose the input tensor $\mbx$ for meta-learner training.
    \item The base model of the meta-learner can now be trained with input $\mbx$ and the corresponding one-hot label $y$.
    \item Training the CNN base model following the few-shot learning described in Algorithm~\ref{alg:MAML}.
    \item The trained few-shot meta-learner can now adapt to new video frames of the unseen scene in a few steps of extra training on small samples.
    \item The trained model can now work in the new scene, where the output of the model is the binary valued prediction of intrusion detection in light of new data streams.
\end{enumerate}

Note that the proposed algorithm can be easily generalised to other small sample problems with similar inter-similarity/intra-similarity issues. For video/image processing applications, one often adopt CNN architecture as the basis of the meta-model. In this paper, the proposed CNN base model consists of 4 convolutional layers, 4 max-pooling layers, 1 linear layers and 1 softmax layers, as shown in the right half of Figure~\ref{fig:basemodel}.

\section{Numerical results}
This section reports the experimental results of the enhanced few-shot learning algorithm in the two datasets. Supportive experiments for the Proposition~\ref{Theorem:1} is firstly presented. The proposed algorithm is then tested against existing algorithms. Performance of different meta-learners with/without engineering tricks are also tested and analysed. All the numerical experiments were implemented by PyTorch in Python 3.
%~\footnote{https://github.com/pytorch/pytorch}.

\subsection{Supportive experiments of Proposition~\ref{Theorem:1}}
\begin{figure}[t!]
\centering
\includegraphics[width=0.99\columnwidth]{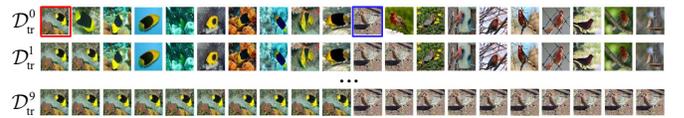}
%\vskip -0.1in
\caption{Illustration of 3 out of the 9 constructed tasks from the MiniImagenet dataset. Each contains 20 images. }
\label{fig:10tasks}
\end{figure}

In this subsection, we assess the Proposition~\ref{Theorem:1} quantitatively on synthetic tasks. As shown in Figure~\ref{fig:10tasks} top panel, we select $20$ images of $2$ classes from MiniImagenet as the comparative task set $\mathcal{D}^{(0)}_{\text{tr}}$. We then try to construct another 9 task sets, i.e., $\mathcal{D}^{(i)}_{\text{tr}}$ for $i=1,\ldots,9$, based on $\mathcal{D}^{(0)}_{\text{tr}}$:
\begin{itemize}
    \item $\mathcal{D}^{(i)}_{\text{tr}}$ for $i=1,\ldots,9$ are initialised by copying $\mathcal{D}^{(0)}_{\text{tr}}$.
    \item 2 out of the 20 images are randomly selected from $\mathcal{D}^{(0)}_{\text{tr}}$, as highlighted by the red and blue frames in the top panel of Figure~\ref{fig:10tasks}.
    \item For the $i$th dataset $\mathcal{D}^{(i)}_{\text{tr}}$, the selected two images are used to replace another $2i$ images in the corresponding dataset. A Gaussian noise ($\sigma^2 = 10$) is added to the replaced images to guarantee all figures are similar but not exactly the same.
\end{itemize}
As a result, the number of same images between $\mathcal{D}^{(0)}_{\text{tr}}$ and $\mathcal{D}^{(i)}_{\text{tr}}$ equals to $20-2i$, which decreases with the index $i$. These task sets are used to update the randomly initialised base model of $f_{\theta}^0$ via (\ref{innerup}), where the gradient descent steps $N$ is set to $5$. Then, we calculate the difference (via MSE and cosine similarity) of the gradients between using $\mathcal{D}^{(0)}_{\text{tr}}$ and using $\mathcal{D}^{(i)}_{\text{tr}}\ (i=1,\ldots,9)$.
%%%%%%%%%%%%%%%%%%%%%%%%%%%%%%%%%%%%%%%%%%%%%%%%%%%%%%%%%%%%%%%%%%%%%%%%%
%%%%%%%%%%%%%%%%%%%%%%%%%%%%%%%%%%%%%%%%%%%%%%%%%%%%%%%%%%%%%%%%%%%%%%%%%

%%%%%%%%%%%%%%%%%%%%%%%%%%%%%%%%%%%%%%%%%%%%%%%%%%%%%%%%%%%%%%%%%%%%%%%%
\begin{figure}[t!]
\centering
\includegraphics[width=0.99\columnwidth]{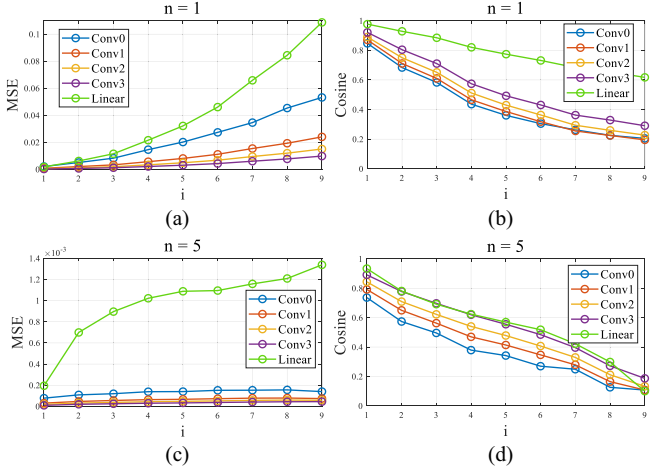}
\caption{Evaluation of the gradient difference between task $\mathcal{D}^{(0)}_{\text{tr}}$ and tasks $\mathcal{D}^{(i)}_{\text{tr}}\ (i=1,\ldots,9)$ under update steps $n=1$ and $n=5$. The parameters of convolutional layers (Conv$0$, Conv$1$, Conv$2$ and Conv$3$) and fully-connected linear layer (Linear) are flattened to vectors. (a) MSE, $n=1$; (b) Cosine, $n=1$;(c) MSE, $n=5$; (d) Cosine, $n=5$.}
%\vskip -0.1in
\label{fig:msecosine}
\end{figure}
%%%%%%%%%%%%%%%%%%%%%%%%%%%%%%%%%%%%%%%%%%%%%%%%%%%%%%%%%%%%%%%%%%%%%%%%%

The MSE and cosine value versus the number of similar images between using $\mathcal{D}^{(0)}_{\text{tr}}$ and $\mathcal{D}^{(i)}_{\text{tr}}\ (i=1,\ldots,9)$ are shown in Figure~\ref{fig:msecosine}, where each curve is averaged over $50$ trials. In Figure~\ref{fig:msecosine}(a), one observes that the MSE between the gradient of each layer increased with index $i$ (or the decrease of the number of same images), and in Figure~\ref{fig:msecosine}(b), one sees that the cosine value decreased with index $i$. These results imply that similar tasks can lead to similar level of gradients, which can easily result in overfitting during model training. Note that (\ref{eq:Theorem1}), (\ref{eq:Theorem2}) and (\ref{eq:Theorem3}) in Proposition~\ref{Theorem:1} theoretically hold only in the simple case with $n=1$. Similar trend can be observed in the Figure~\ref{fig:msecosine}(c) and (d), where the base model in each task performs multiple gradient descent steps ($n>1$). Although we cannot provide the derivations for a general CNN with multiple update steps, one could draw the same conclusion according the experiments.

%%%%%%%%%%%%%%%%%%%%%%%%%%%%%%%%%%%%%%%%%%%%%%%%%%%%%%%%%%%%%%%%%%%%%%%%%%%%%%%%%%%%%%%
\begin{figure}[t!]
\centering
\includegraphics[width=0.99\columnwidth]{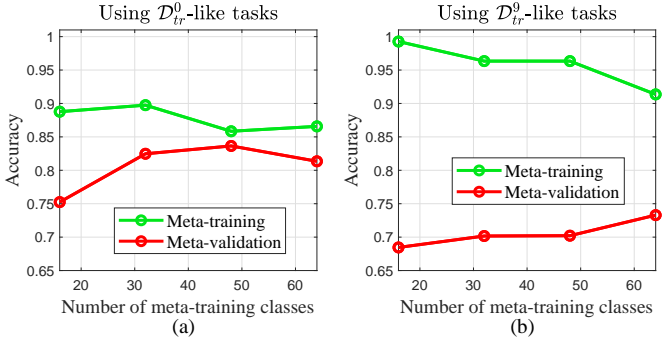}
%\vskip -0.1in
\caption{Learning curves of $f_{\theta}^0$ using different constructed tasks from MiniImagenet. (a) MAML using $\mathcal{D}^{(0)}_{\text{tr}}$-like tasks. (b) MAML using $\mathcal{D}^{(9)}_{\text{tr}}$-like tasks.}
\label{fig:D0D9}
\end{figure}
%%%%%%%%%%%%%%%%%%%%%%%%%%%%%%%%%%%%%%%%%%%%%%%%%%%%%%%%%%%%%%%%%%%%%%%%%%%%%%%%%%%%%%%%

To demonstrate that similar training scenarios will cause overfitting, in Figure~\ref{fig:D0D9}, we further compare the simplified learning curves of original MAML in~\cite{MAML} using different constructed tasks from MiniImagenet, where the greater gap of accuracy between training and verification means the more severe overfitting of the meta model. In Figure~\ref{fig:D0D9} (a), we can observe the overfitting is improved as the increasing of the number of training classes using $\mathcal{D}^{(0)}_{\text{tr}}$-like tasks. However, when the number of meta-training classes is up to the maximum of $64$, MAML still suffers from the severe overfitting by using the synthetic $\mathcal{D}^{(9)}_{\text{tr}}$-like tasks as shown in Figure~\ref{fig:D0D9} (b), which supports our statement that similar training scenarios cause MAML overfitting.

\subsection{Ablation study of the improved base model}
%%%%%%%%%%%%%%%%%%%%%%%%%%%%%%%%%%%%%%%%%%%%%%%%%%%%%%%%%%%%%%%%%%%%%%%%%%%%%%%%%%%%%%%
\begin{figure}[t!]
\centering
\includegraphics[width=0.99\columnwidth]{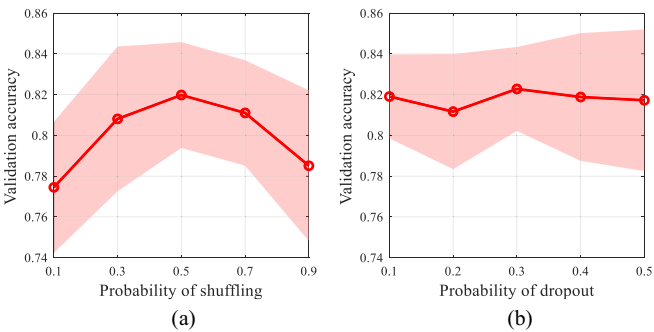}
%\vskip -0.1in
\caption{The $10$-fold cross-validation results of accuracy and variance for tuning the hyper-parameters. (a) Results with different probabilities of shuffling; (b) Results with different probabilities of dropout.}
\label{fig:tune}
%\vskip -0.1in
\end{figure}
%%%%%%%%%%%%%%%%%%%%%%%%%%%%%%%%%%%%%%%%%%%%%%%%%%%%%%%%%%%%%%%%%%%%%%%%%%%%%%%%%%%%%%%%

Here we evaluate the proposed measures for improving the base model, involving shuffling label and neurons dropout. Following the settings on MiniImagenet in~\cite{MAML}, for $K$-shot intrusion detection, we set support set size $2K = 10$, query set size $2Q = 30$, meta-batch size $I = 2$,  inner-loop learning rate $\alpha = 0.01$, meta-learning rate $\beta = 0.001$ and maximum iteration number $J = 60000$. The model is trained using $N = 5$ gradient steps and evaluated using $10$ gradient steps. In the meta-training phase, the base model is evaluated per $200$ iterations and early stopped when the accuracy no longer increases after $2000$ iterations. These parameter settings are used in all following experiments unless otherwise specified.

%%%%%%%%%%%%%%%%%%%%%%%%%%%%%%%%%%%%%%%%%%%%%%%%%%%%%%%%%%%%%%%%%%%%%%%%%%%%%%%%%%%%%%%
\begin{figure*}[t!]
\centering
\includegraphics[width=1.8\columnwidth]{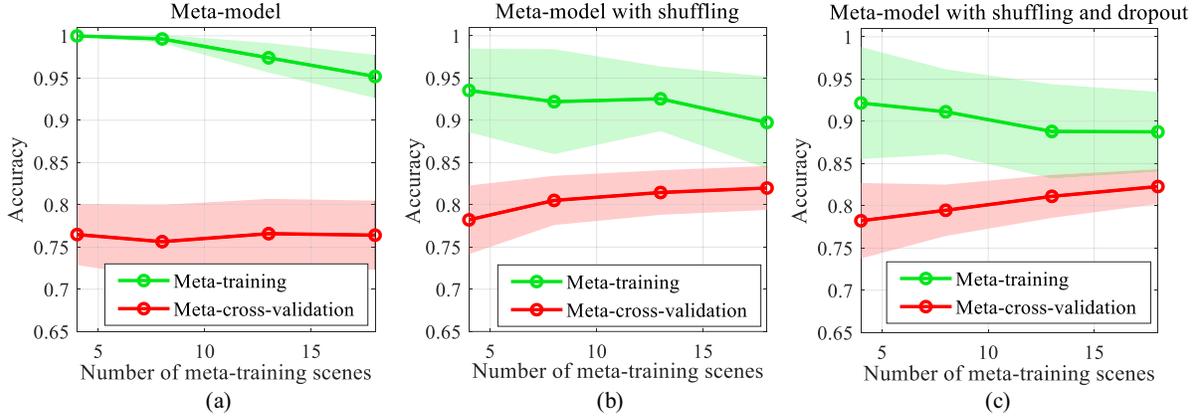}
%\vskip -0.1in
\caption{Learning curves of different base models. (a) MAML of $f_{\theta}^1$; (b) MAML of $f_{\theta}^1$ with shuffling label scheme; (c) MAML of $f_{\theta}^1$ with both shuffling label and neuron dropout schemes.}
\label{fig:learningcurve}
%\vskip -0.1in
\end{figure*}
%%%%%%%%%%%%%%%%%%%%%%%%%%%%%%%%%%%%%%%%%%%%%%%%%%%%%%%%%%%%%%%%%%%%%%%%%%%%%%%%%%%%%%%%

\par
There are two hyper-parameters need to be tuned involving the probability of shuffling and probability of dropout. After drawing a specific task from the local dataset, we need to decide with what probability the intrusive images are labelled by $[1,0]$, which is denoted as the probability of shuffling. Although it is treated as $0.5$, it should to be checked due to the imbalance of positive and negative samples. In our model $f_{\theta}^1$, we carry the $10$-fold cross-validation using $18$ meta-training scenes and $2$ meta-cross-validation scenes. The mean validation accuracy versus probability of shuffling is shown in Figure~\ref{fig:tune}(a), where each trained model adapts to $50$ tasks drawn from $2$ meta-cross-validation scenes. We can observe that the highest accuracy performance is achieved at the probability of shuffling $0.5$. Further, we dropout neurons in linear layer with different probabilities and report the cross-validation results in Figure~\ref{fig:tune}(b). We can observe that the high probability of dropout lead to the high variance of accuracy. We set probability of dropout as $0.3$ owing the high accuracy and low variance.

\par
We use the learning curves to find out whether the base model suffers more from a variance error or a bias error with the increase on the number of meta-training scenes. It helps to check out the overfitting for different base models. Three base models are evaluated involving our model $f_{\theta}^1$, $f_{\theta}^1$ with shuffling label and $f_{\theta}^1$ with shuffling label and neuron dropout for the linear layer. We perform $10$-fold cross-validation with different numbers of meta-training scenes and report the mean validation accuracy in Figure~\ref{fig:learningcurve}. As shown in Figure~\ref{fig:learningcurve}(a), we can observe the high bias between the accuracy of training and cross-validation, which means that $f_{\theta}^1$ suffers severe overfitting.
By using shuffling label, the overfitting of $f_{\theta}^1$ is improved due to the lower bias and smaller variance in Figure~\ref{fig:learningcurve}(b). In Figure~\ref{fig:learningcurve}(b) and Figure~\ref{fig:learningcurve}(c), the best accuracies of cross-validation without and with dropout are $81.98\%\pm2.60\%$ and $82.28\%\pm2.06\%$, respectively, and the subtle difference also implies that adding dropout layer leads to better accuracy and smaller variance. According to Figure~\ref{fig:learningcurve}(c), we can observe that the final base model will probably not benefit much from more training scenes.

% 改到这了

\subsection{Ablation study of the additional mask input}
\begin{figure}[t!]
\centering
\includegraphics[width=0.99\columnwidth]{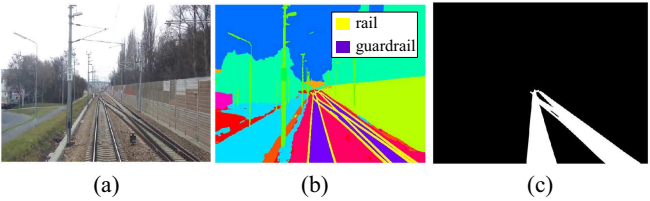}
%\vskip -0.1in
\caption{An illustration of the pre-processing of the RailSem19 sample. (a) Original image; (b) Original mask; (c) Mask after selecting the label of \emph{rail} and \emph{guardrail}.}
\label{fig:rs}
\end{figure}

In this paper, we train the FCNs using RailSem19\footnote{https://wilddash.cc/}~\cite{Zendel_2019} which is a dataset for semantic rail scene understanding with dense label masks. The RailSem19 has $8500$ images taken from the ego-perspective of rail vehicles. As shown in Figure~\ref{fig:rs}, we select the pixel labels of \emph{rail} and \emph{guardrail}, which are the bases to the rail track wheels, as segmented objects, and treat the other pixels as the background. After removing the images without these two labels, there are $7772$ images left, in which $6997$ ($90\%$) images are used for training and $776$ ($10\%$) images are used for test. In addition,
all the images are resize to $640\times 480$.

%%%%%%%%%%%%%%%%%%%%%%%%%%%%%%%%%%%%%%%%%%%%%%%%%%%%%%%%%%%%%%%%%%%%%%%%%%%%%%%%%%%%%%%
%%%%%%%%%%%%%%%%%%%%%%%%%%%%%%%%%%%%%%%%%%%%%%%%%%%%%%%%%%%%%%%%%%%%%%%%%%%%%%%%%%%%%%%%

%%%%%%%%%%%%%%%%%%%%%%%%%%%%%%%%%%%%%%%%%%%%%%%%%%%%%%%%%%%%%%%%%%%%%%%%%%%%%%%%%%%%%%%
\begin{figure}[t!]
\centering
\includegraphics[width=0.99\columnwidth]{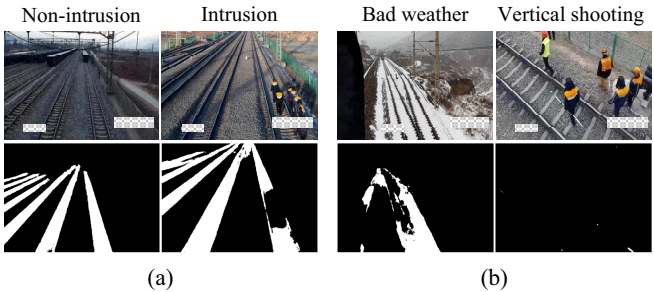}
%\vskip -0.1in
\caption{An illustration of input images (top) in the railway dataset and output masks (bottom) of the FCN-8s. (a) Results of non-intrusive and intrusive images; (a) Failed extractions in the cases of bad weather and vertical shooting.}
\label{fig:FCNre}
%\vskip -0.2in
\end{figure}
%%%%%%%%%%%%%%%%%%%%%%%%%%%%%%%%%%%%%%%%%%%%%%%%%%%%%%%%%%%%%%%%%%%%%%%%%%%%%%%%%%%%%%%%

\par
The backbone network of FCNs in our experiments was VGG16~\cite{simonyan2014very} pre-trained on ImageNet. We train FCN-32s, FCN-16s and FCN-8s, which combine different coarse low layer information with fine higher layer information, by stochastic gradient descent (SGD) with momentum. We set momentum to $0.7$, initial learning rate to $0.01$, mini-batch size of images to $4$ and maximum number of epoches to $50$.
Two popular criteria are used to evaluate the segmentation performance involving pixel accuracy (PA) defined by $\frac{L\cap R}{L}$ and intersection over union (IU) defined by $\frac{L\cap R}{L\cup R}$, where $L$ denotes the set of the ground truth and $R$ denotes the set of segmentation result. Segmented masks with higher PA and IU are considered with better quality. The test performances of the FCNs are reported in Table~\ref{tab:FCN}, where the FCN-8s leads to the best IU and PA. We illustrate some segmented results of local image using the trained FCN-8s in Figure~\ref{fig:FCNre}.
As shown in Figure~\ref{fig:FCNre}(a), we can observe that the track area can be located by the FCN and the intrusions affect the result of segmentation. However, there are also bad segmented results in the cases which RailSem19 dataset does not covers, as shown in Figure~\ref{fig:FCNre}(b), track are hidden under snow or the image are shot in vertical angle. In this paper, the FCN-8s is selected as the segmenter for extracting track area.
%%%%%%%%%%%%%%%%%%%%%%%%%%%%%%%%%%%%%%%%%%%%%%%%%%%%%%%%%%%%%%%%%%%%%%%%%%
\begin{table}[t!]
\begin{center}
\caption{The averaged segmentation performance of FCNs over the test trails from RailSem19.}
%\vskip 0.1in
\label{tab:FCN}
\begin{tabular}{cccc}
\toprule
Index      & FCN-32s&FCN-16s& FCN-8s \\
\midrule
Mean PA & $95.49\%$&$96.54\%$&$96.70\%$\\

Mean IU & $90.66\%$&$92.76\%$&$93.08\%$\\
\bottomrule
\end{tabular}
\end{center}
\vskip -0.1in
\end{table}
%%%%%%%%%%%%%%%%%%%%%%%%%%%%%%%%%%%%%%%%%%%%%%%%%%%%%%%%%%%%%%%%%%%%%%%%%%%%%%%%

%%%%%%%%%%%%%%%%%%%%%%%%%%%%%%%%%%%%%%%%%%%%%%%%%%%%%%%%%%%%%%%%%%%%%%%%%%%%%%%%%%%%%%%%
%\begin{figure}[t!]
%\centering
%\includegraphics[width=0.95\columnwidth]{fig14}
%\vskip -0.1in
%\caption{The test performance of the proposed intrusion detection method with and without additional mask.}
%\vskip -0.1in
%\label{fig:vsnomask}
%\end{figure}
%%%%%%%%%%%%%%%%%%%%%%%%%%%%%%%%%%%%%%%%%%%%%%%%%%%%%%%%%%%%%%%%%%%%%%%%%%%%%%%%%%%%%%%%%

%%%%%%%%%%%%%%%%%%%%%%%%%%%%%%%%%%%%%%%%%%%%%%%%%%%%%%%%%%%%%%%%%%%%%%%%%%%%%%%%%%%%%%%%%
\renewcommand{\arraystretch}{1.2}
\begin{table*}[t!]
\begin{center}
\caption{The test performances of the proposed $K$-shot intrusion detection algorithm with and without masks.}
%\vskip 0.1in
%\begin{small}
\label{tab:vsnomask}
\begin{tabular}{ccccccccccc}
\toprule
\multicolumn{1}{c}{\multirow{2}{*}{Index}} & \multicolumn{5}{c}{Without mask} & \multicolumn{5}{c}{With mask} \\
\multicolumn{1}{c}{} &K=$1$&K=$2$&K=$3$&K=$4$&K=$5$&K=$1$&K=$2$&K=$3$&K=$4$&K=$5$ \\
\midrule
FPR
&$20.27\%$&$16.00\%$&$14.73\%$&$13.4\%$&$9.33\%$&$15.60\%$&$12.33\%$&$12.33\%$&$7.60\%$&$6.80\%$\\
FNR
&$18.47\%$&$16.27\%$&$12.47\%$&$9.60\%$&$8.20\%$&$20.40\%$&$13.53\%$&$12.53\%$&$12.40\%$&$9.40\%$\\
Accuracy
&$61.27\%$&$67.73\%$&$72.80\%$&$76.99\%$&$82.47\%$&$64.00\%$&$74.13\%$&$75.13\%$&$80.00\%$&$83.80\%$\\
Processing speed & \multicolumn{5}{c}{0.34 ms/img} & \multicolumn{5}{c}{33.91 ms/img (FCN-8s)+0.35 ms/img} \\
\bottomrule
\end{tabular}
%\end{small}
\end{center}
\vskip -0.2in
\end{table*}
%%%%%%%%%%%%%%%%%%%%%%%%%%%%%%%%%%%%%%%%%%%%%%%%%%%%%%%%%%%%%%%%%%%%%%%%%%%%%%%%%%%%%%%%%%%%

We use the division pattern of training scenes and validation scenes in the local dataset that leads to the closest accuracy to the average performance of the base model in the cross-validation experiments. Note that we do not choose the best one due to it may lead to high bias between all training scenes and two test scenes that never been seen before. We drawn $50$ tasks from test scenes each of which has $K$ intrusion samples and $K$ non-intrusion samples. The average test performances involve false-positive ratio (FPR, i.e., false alarm rate), false-negative ratio (FNR, i.e., missing rate), accuracy and speed of all trained base models with $K = 1,\ldots,5$ are reported in Table~\ref{tab:vsnomask}, where the processing speed is evaluated on $1$ GTX1080Ti GPU. We can observe that using the additional mask input leads to the better detection accuracy in different cases because it produces to the smaller FPR. Note that the detection process using the additional mask input is divided into two parts: segmentation and detection, and we can observe that segmentation by FCN-$8$s is time consuming due to using more network parameters compared with detection.

\subsection{Comparison with classical supervised learning method}
%%%%%%%%%%%%%%%%%%%%%%%%%%%%%%%%%%%%%%%%%%%%%%%%%%%%%%%%%%%%%%%%%%%%%%%%%%%%%%%%%%%%%%%
\begin{figure}[t!]
\centering
\includegraphics[width=0.99\columnwidth]{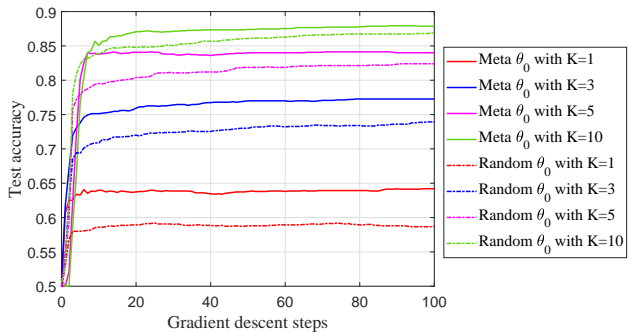}
\vskip -0.1in
\caption{The test performances of the base model with random initial conditions and meta initial conditions.}
\vskip -0.1in
\label{fig:vsrandom}
\end{figure}
%%%%%%%%%%%%%%%%%%%%%%%%%%%%%%%%%%%%%%%%%%%%%%%%%%%%%%%%%%%%%%%%%%%%%%%%%%%%%%%%%%%%%%%%

Note that the benchmark of the proposed meta-learning intrusion detection method is the classical supervised learning method. The classical supervised learning initialises the network randomly but the proposed method of meta-learning initialises with the meta-trained $\theta_0$. In this subsection, we discuss the advantages of the proposed meta-learning method. We evaluate both two kinds of model with additional mask on $50$ tasks drawn from the two test railway scenes in the local dataset. Then we train for each tasks in the two case with the same settings for fair comparison except the initial condition. The mean test accuracy versus the number of gradient descent steps is shown in Figure~\ref{fig:vsrandom}. Note that the meta initial conditions are meta-trained with updated steps $N=5$ and here we evaluate them with $N=100$ for comparison. It can be seen that learning with the meta initialisation adapts fast with a few number of gradient descent steps while learning with random initialisation converges more slowly.
Moreover, we can observe that the meta initial condition leads to $4.8\%$, $3.9\%$, $3.7\%$ and $2.3\%$ average gains after $20$ steps with the shot of $K=1$, $K=3$, $K=5$ and $K=10$, respectively, compared with the traditional random initialisation.
This means the proposed meta learning method is more economical owing to the high accuracy using a few number of training steps when it is applied on railway scenes, as for the randomly initialised CNNs, only by using more training samples can it lead to the better performance than meta model.

%%%%%%%%%%%%%%%%%%%%%%%%%%%%%%%%%%%%%%%%%%%%%%%%%%%%%%%%%%%%%%%%%%%%%%%%%%%%%%%%%%%%%%%%%
\begin{table}[t!]
\begin{center}
\caption{The $5$-shot test performances of the proposed algorithm and competing ProtoNets for intrusion detection with different techniques, where ``Ori'' denotes the original version.}
%\vskip 0.1in
%\begin{small}
\label{tab:pnvsmaml}
\begin{tabular}{cccccc}
\toprule
\multicolumn{4}{c}{Technique} & \multicolumn{2}{c}{Method} \\
Ori         & +shuffling   & +dropout  & +mask    & ProtoNets & proposed     \\
\midrule
\checkmark &             &           &           &$78.27\%$ &$78.80\%$   \\
\checkmark &\checkmark   &           &           &$78.53\%$ &$80.41\%$   \\
\checkmark &\checkmark   &           &\checkmark &$79.33\%$ &$82.60\%$   \\
\checkmark &\checkmark   &\checkmark &\checkmark & ---      &$83.80\%$ \\
\bottomrule
\end{tabular}
%\end{small}
\end{center}
\vskip -0.2in
\end{table}
%%%%%%%%%%%%%%%%%%%%%%%%%%%%%%%%%%%%%%%%%%%%%%%%%%%%%%%%%%%%%%%%%%%%%%%%%%%%%%%%%%%%%%%%%%%%

\subsection{Comparison with metric-based meta-learning method}
In this subsection, we evaluate the detection performance of proposed method compared with another state-of-the-art meta-learning pipeline, i.e. metric-based meta-learning,
of which the prototypical networks (ProtoNets) proposed in \cite{NIPS2017_6996} is a representative method. In the meta-training phase, ProtoNets learn the metric space in which classification can be performed by computing distances to prototype centers of intrusion and non-intrusion images for each few-shot task, once the metric space is learned, they can be directly used to detect intrusions and do not require the fine-tuning process.
\par
Following the settings in \cite{NIPS2017_6996}, we use the Euclidean distance and the SGD with Adam~\cite{KingmaB14} to train all ProtoNets. The learning rate is initially set to $5\times10^{-4}$ and cut in half every $2000$ meta-training tasks, and the same early-stop strategy is applied to the two algorithm. In Table~\ref{tab:pnvsmaml}, we report the average $5$-shot detection accuracy of $50$ meta-test tasks with different techniques used in this paper. Note that there is no linear layer in ProtoNets so the dropout cannot be used. It can be observed that the proposed detection method based on MAML outperforms ProtoNets in all cases.

\section{Conclusion}
This paper presents an enhanced few-shot learning solution for railway video intrusion detection with small samples. The railway video surveillance data suffers the low inter-similarity and high intra-similarity issues, which was addressed by the proposed algorithm. As far as we aware, this is the first time the few-shot learning algorithm applied to the small sample railway video intrusion dection problems. Numerical results demonstrate that the proposed method outperforms existing methods and achieve satisfactory results in terms of prediction accuracy and algorithm efficiency. Upon successful execution of the proposed algorithm, the trained meta-model can quickly adapt to an unseen railway scene with only a few new video frame samples (with a few number of gradient descent steps). The proposed method provides an economically and computationally efficient alternative to the railway video intrusion detection applications.

\appendices

%\section{}

%% use section* for acknowledgment
%\section*{Acknowledgment}

\ifCLASSOPTIONcaptionsoff
  \newpage
\fi



%\bibliographystyle{IEEEtran}
%\begin{thebibliography}{1}
%\end{thebibliography}

\bibliographystyle{IEEEtran}
\bibliography{IEEEabrv,mybibfile}

\begin{thebibliography}{10}
\providecommand{\url}[1]{#1}
\csname url@samestyle\endcsname
\providecommand{\newblock}{\relax}
\providecommand{\bibinfo}[2]{#2}
\providecommand{\BIBentrySTDinterwordspacing}{\spaceskip=0pt\relax}
\providecommand{\BIBentryALTinterwordstretchfactor}{4}
\providecommand{\BIBentryALTinterwordspacing}{\spaceskip=\fontdimen2\font plus
\BIBentryALTinterwordstretchfactor\fontdimen3\font minus
  \fontdimen4\font\relax}
\providecommand{\BIBforeignlanguage}[2]{{%
\expandafter\ifx\csname l@#1\endcsname\relax
\typeout{** WARNING: IEEEtran.bst: No hyphenation pattern has been}%
\typeout{** loaded for the language `#1'. Using the pattern for}%
\typeout{** the default language instead.}%
\else
\language=\csname l@#1\endcsname
\fi
#2}}
\providecommand{\BIBdecl}{\relax}
\BIBdecl

\bibitem{4730859}
T.~S.~K. {Chan} and K.~S.~M. {Chung}, ``Applications and selections of
  intelligent surveillance system in railway industry,'' in \emph{International
  Conference on Railway Engineering - Challenges for Railway Transportation in
  Information Age}, March 2008, pp. 1--6.

\bibitem{101177}
Z.~Xie and Y.~Qin, ``High-speed railway perimeter intrusion detection approach
  based on internet of things,'' \emph{Advances in Mechanical Engineering},
  vol.~11, no.~2, p. 1687814018821511, 2019.

\bibitem{guo2012intrusion}
B.~Guo, L.~Zhu, and H.~Shi, ``Intrusion detection algorithm for railway
  clearance with rapid dbscan clustering,'' \emph{Chin. J. Sci. Instrum},
  vol.~33, pp. 241--247, 2012.

\bibitem{CATALANO201791}
A.~Catalano, F.~A. Bruno, C.~Galliano, M.~Pisco, G.~V. Persiano, A.~Cutolo, and
  A.~Cusano, ``An optical fiber intrusion detection system for railway
  security,'' \emph{Sensors and Actuators A: Physical}, vol. 253, pp. 91 --
  100, 2017.

\bibitem{4692395}
S.~{Oh}, G.~{Kim}, and {Hanmin Lee}, ``A monitoring system with ubiquitous
  sensors for passenger safety in railway platform,'' in \emph{Internatonal
  Conference on Power Electronics}, Oct 2007, pp. 289--294.

\bibitem{luy2018initial}
M.~L{\"u}y, E.~{\c{C}}am, F.~Ulam{\i}{\c{s}}, I.~Uzun, and S.~{\.I}. Ak{\i}n,
  ``Initial results of testing a multilayer laser scanner in a collision
  avoidance system for light rail vehicles,'' \emph{Applied Sciences}, vol.~8,
  no.~4, p. 475, 2018.

\bibitem{pu2014study}
Y.-R. Pu, L.-W. Chen, and S.-H. Lee, ``Study of moving obstacle detection at
  railway crossing by machine vision,'' \emph{Information Technology Journal},
  vol.~13, no.~16, pp. 2611--2618, 2014.

\bibitem{NIPS2012_4824}
A.~Krizhevsky, I.~Sutskever, and G.~E. Hinton, ``Imagenet classification with
  deep convolutional neural networks,'' in \emph{Advances in neural information
  processing systems}, 2012, pp. 1097--1105.

\bibitem{MAML}
C.~Finn, P.~Abbeel, and S.~Levine, ``Model-agnostic meta-learning for fast
  adaptation of deep networks,'' in \emph{International Conference on Machine
  Learning (ICML)}, 2017, pp. 1126--1135.

\bibitem{7478072}
E.~{Shelhamer}, J.~{Long}, and T.~{Darrell}, ``Fully convolutional networks for
  semantic segmentation,'' \emph{IEEE Transactions on Pattern Analysis and
  Machine Intelligence}, vol.~39, no.~4, pp. 640--651, April 2017.

\bibitem{1336499}
J.~{Vazquez}, M.~{Mazo}, J.~L. {Lazaro}, C.~A. {Luna}, J.~{Urena}, J.~J.
  {Garcia}, J.~{Cabello}, and L.~{Hierrezuelo}, ``Detection of moving objects
  in railway using vision,'' in \emph{IEEE Intelligent Vehicles Symposium,
  2004}, June 2004, pp. 872--875.

\bibitem{Shi2015Study}
H.~M. Shi, H.~Chai, Y.~Wang, and Z.~J. Yu, ``Study on railway embedded
  detection algorithm for railway intrusion based on object recognition and
  tracking,'' \emph{Journal of the China Railway Society}, vol.~37, no.~7, pp.
  58--65, 2015.

\bibitem{Guo2016High}
B.~Guo, L.~Yang, H.~Shi, Y.~Wang, and X.~Xu, ``High-speed railway clearance
  intrusion detection algorithm with fast background subtraction,''
  \emph{Chinese Journal of Scientific Instrument}, 2016.

\bibitem{nakasone2017frontal}
R.~Nakasone, N.~Nagamine, M.~Ukai, H.~Mukojima, D.~Deguchi, and H.~Murase,
  ``Frontal obstacle detection using background subtraction and frame
  registration,'' \emph{Quarterly Report of Rtri}, vol.~58, no.~4, pp.
  298--302, 2017.

\bibitem{7506117}
X.~{Gibert}, V.~M. {Patel}, and R.~{Chellappa}, ``Deep multitask learning for
  railway track inspection,'' \emph{IEEE Transactions on Intelligent
  Transportation Systems}, vol.~18, no.~1, pp. 153--164, Jan 2017.

\bibitem{8006280}
G.~{Krummenacher}, C.~S. {Ong}, S.~{Koller}, S.~{Kobayashi}, and J.~M.
  {Buhmann}, ``Wheel defect detection with machine learning,'' \emph{IEEE
  Transactions on Intelligent Transportation Systems}, vol.~19, no.~4, pp.
  1176--1187, April 2018.

\bibitem{8516370}
G.~{Kang}, S.~{Gao}, L.~{Yu}, and D.~{Zhang}, ``Deep architecture for
  high-speed railway insulator surface defect detection: Denoising autoencoder
  with multitask learning,'' \emph{IEEE Transactions on Instrumentation and
  Measurement}, vol.~68, no.~8, pp. 2679--2690, Aug 2019.

\bibitem{wang2019adaptive}
Y.~Wang, L.~Zhu, Z.~Yu, and B.~Guo, ``An adaptive track segmentation algorithm
  for a railway intrusion detection system,'' \emph{Sensors}, vol.~19, no.~11,
  p. 2594, 2019.

\bibitem{8832957}
H.~{Huang}, L.~{Liang}, G.~{Zhao}, Y.~{Yang}, and K.~{Ou}, ``Railway clearance
  intrusion detection in aerial video based on convolutional neural network,''
  in \emph{Chinese Control And Decision Conference (CCDC)}, June 2019, pp.
  1644--1648.

\bibitem{simonyan2014very}
K.~Simonyan and A.~Zisserman, ``Very deep convolutional networks for
  large-scale image recognition,'' \emph{arXiv preprint arXiv:1409.1556}, 2014.

\bibitem{guo2019high}
B.~Guo, J.~Shi, L.~Zhu, and Z.~Yu, ``High-speed railway clearance intrusion
  detection with improved ssd network,'' \emph{Applied Sciences}, vol.~9,
  no.~15, p. 2981, 2019.

\bibitem{liu2016ssd}
W.~Liu, D.~Anguelov, D.~Erhan, C.~Szegedy, S.~Reed, C.-Y. Fu, and A.~C. Berg,
  ``Ssd: Single shot multibox detector,'' in \emph{European conference on
  computer vision (ECCV)}.\hskip 1em plus 0.5em minus 0.4em\relax Springer,
  2016, pp. 21--37.

\bibitem{7485869}
S.~{Ren}, K.~{He}, R.~{Girshick}, and J.~{Sun}, ``Faster r-cnn: Towards
  real-time object detection with region proposal networks,'' \emph{IEEE
  Transactions on Pattern Analysis and Machine Intelligence}, vol.~39, no.~6,
  pp. 1137--1149, June 2017.

\bibitem{8978612}
T.~{Ye}, X.~{Zhang}, Y.~{Zhang}, and J.~{Liu}, ``Railway traffic object
  detection using differential feature fusion convolution neural network,''
  \emph{IEEE Transactions on Intelligent Transportation Systems}, pp. 1--13,
  2020.

\bibitem{schmidhuber}
J.~Schmidhuber, ``Evolutionary principles in self-referential learning. on
  learning now to learn: The meta-meta-meta...-hook,'' Diploma Thesis,
  Technische Universitat Munchen, Germany, 14 May 1987.

\bibitem{NIPS2016_6385}
O.~Vinyals, C.~Blundell, T.~Lillicrap, D.~Wierstra \emph{et~al.}, ``Matching
  networks for one shot learning,'' in \emph{Advances in neural information
  processing systems}, 2016, pp. 3630--3638.

\bibitem{NIPS2017_6996}
J.~Snell, K.~Swersky, and R.~Zemel, ``Prototypical networks for few-shot
  learning,'' in \emph{Advances in neural information processing systems},
  2017, pp. 4077--4087.

\bibitem{Sung_2018_CVPR}
F.~Sung, Y.~Yang, L.~Zhang, T.~Xiang, P.~H. Torr, and T.~M. Hospedales,
  ``Learning to compare: Relation network for few-shot learning,'' in
  \emph{IEEE Conference on Computer Vision and Pattern Recognition (CVPR)},
  June 2018.

\bibitem{RaviL17}
S.~Ravi and H.~Larochelle, ``Optimization as a model for few-shot learning,''
  in \emph{International Conference on Learning Representations (ICLR)}.\hskip
  1em plus 0.5em minus 0.4em\relax OpenReview.net, 2017.

\bibitem{wu19d}
Y.~Wu, M.~Rosca, and T.~Lillicrap, ``Deep compressed sensing,'' in
  \emph{International Conference on Machine Learning (ICML)}, vol.~97, 2019,
  pp. 6850--6860.

\bibitem{8815426}
S.~{Park}, H.~{Jang}, O.~{Simeone}, and J.~{Kang}, ``Learning how to demodulate
  from few pilots via meta-learning,'' in \emph{International Workshop on
  Signal Processing Advances in Wireless Communications (SPAWC)}, July 2019,
  pp. 1--5.

\bibitem{8761319}
H.~{Mao}, H.~{Lu}, Y.~{Lu}, and D.~{Zhu}, ``Roemnet: Robust meta learning based
  channel estimation in ofdm systems,'' in \emph{IEEE International Conference
  on Communications (ICC)}, May 2019, pp. 1--6.

\bibitem{Yin20}
M.~Yin, G.~Tucker, M.~Zhou, S.~Levine, and C.~Finn, ``Meta-learning without
  memorization,'' in \emph{International Conference on Learning Representations
  (ICLR)}.\hskip 1em plus 0.5em minus 0.4em\relax OpenReview.net, 2020.

\bibitem{SrivastavaHKSS14}
N.~Srivastava, G.~Hinton, A.~Krizhevsky, I.~Sutskever, and R.~Salakhutdinov,
  ``Dropout: a simple way to prevent neural networks from overfitting,''
  \emph{The journal of machine learning research}, vol.~15, no.~1, pp.
  1929--1958, 2014.

\bibitem{ioffe2015batch}
S.~Ioffe and C.~Szegedy, ``Batch normalization: Accelerating deep network
  training by reducing internal covariate shift,'' \emph{arXiv preprint
  arXiv:1502.03167}, 2015.

\bibitem{Zendel_2019}
O.~Zendel, M.~Murschitz, M.~Zeilinger, D.~Steininger, S.~Abbasi, and
  C.~Beleznai, ``Railsem19: A dataset for semantic rail scene understanding,''
  in \emph{IEEE Conference on Computer Vision and Pattern Recognition (CVPR)
  Workshops}, June 2019.

\bibitem{KingmaB14}
D.~P. Kingma and J.~Ba, ``Adam: {A} method for stochastic optimization,'' in
  \emph{International Conference on Learning Representations (ICLR)}.\hskip 1em
  plus 0.5em minus 0.4em\relax OpenReview.net, 2015.

\end{thebibliography}

\end{document}